\documentclass[authoryear,final,1p,12pt,times]{elsarticle}

\usepackage{amssymb}
\usepackage{booktabs}
\usepackage{lineno}
\usepackage{subcaption}
\usepackage{url}

\journal{journal}

\usepackage{boxedminipage}
\usepackage{textpos}

\begin{document}

\begin{frontmatter}

\title{GANmapper: geographical data translation}
          
\author[doa]{Abraham Noah Wu}
\ead{abrahamwu@u.nus.edu}
\author[doa,dre]{Filip Biljecki\corref{cor1}}
\ead{filip@nus.edu.sg}

\address[doa]{Department of Architecture, National University of Singapore, Singapore}
\address[dre]{Department of Real Estate, National University of Singapore, Singapore}

\cortext[cor1]{Corresponding author}

\begin{abstract}
We present a new method to create spatial data using a generative adversarial network (GAN).
Our contribution uses coarse and widely available geospatial data to create maps of less available features at the finer scale in the built environment, bypassing their traditional acquisition techniques (e.g.\ satellite imagery or land surveying).
In the work, we employ land use data and road networks as input to generate building footprints and conduct experiments in 9 cities around the world.
The method, which we implement in a tool we release openly, enables the translation of one geospatial dataset to another with high fidelity and morphological accuracy. 
It may be especially useful in locations missing detailed and high-resolution data and those that are mapped with uncertain or heterogeneous quality, such as much of OpenStreetMap.
The quality of the results is influenced by the urban form and scale.
In most cases, the experiments suggest promising performance as the method tends to truthfully indicate the locations, amount, and shape of buildings.
The work has the potential to support several applications, such as energy, climate, and urban morphology studies in areas previously lacking required data or inpainting geospatial data in regions with incomplete data.

\end{abstract}


\end{frontmatter}

\begin{textblock*}{\textwidth}(0cm,-20.1cm)
\begin{center}
\begin{footnotesize}
\hspace*{-2.8cm}
\begin{boxedminipage}{1.4\textwidth}
This is an Accepted Manuscript of an article published by Taylor \& Francis in the \emph{International Journal of Geographical Information Science} in 2022, available online: \url{https://doi.org/10.1080/13658816.2022.2041643}.
Cite as:\\
Wu AN, Biljecki F (2022): GANmapper: geographical data translation. \textit{International Journal of Geographical Information Science}.\\
It is deposited under the terms of the Creative Commons Attribution-NonCommercial License (\url{http://creativecommons.org/licenses/by-nc/4.0/}), which permits non-commercial re-use, distribution, and reproduction in any medium, provided the original work is properly cited.
\end{boxedminipage}
\end{footnotesize}
\end{center}
\end{textblock*}

\section{Introduction}\label{sc:introduction}

Generative adversarial networks (GANs) are a type of generative models introduced by \citet{goodfellow2014generative}, which have rapidly gained currency in a variety of application domains, such as thermal comfort, energy, and design \citep{Quintana.2020,Yan.2020,Rachele.2021}.
Using a generator-discriminator model pair in the training process, the generator in a GAN gradually learns to create data distributions that pass the checks by the discriminator, therefore producing patterns that closely resemble the original dataset.

With sufficient training data, state-of-the-art GANs are able to generate synthetic photo-realistic images that can deceive the human eye \citep{brock2018large, karras2019style, karras2020analyzing}.
In certain GAN architectures, the input data provides constraints or contextual information that controls the output of the generator. This technique can be applied on different data pairs and makes the output more controllable. For example, users can draw simple color-coded shapes as input masks and the generator will be able to create photo-realistic images according to the elements given by the user \citep{park2019gaugan} . 
In other explorations, a series of texts that describes a scene can be used as input and the generator is also able to generate eye-deceiving images according to the input information \citep{pmlr1, Xu_2018_CVPR}.

Currently, many explorations of image data translation with GANs focus more on cartographic style transfer such as translating satellite images to electronic maps in the style of OpenStreetMap or Google Maps and vice versa \citep{li2020mapgan, 9200630, ganguli2019geogan, 9205243}.
While these explorations undoubtedly raised the interest of GANs in GIS, their scientific application is limited since feature extraction algorithms could already extract geographical features from satellite maps into vectors with high efficiency and accuracy \citep{wu2020roofpedia, WALDNER2020111741}.

In the other fields, GANs has already found many scientific and practical uses.
One notable use is to generate synthetic data for deep learning training. Instead of generating synthetic examples manually, GANs can be used to generate realistic-looking synthetic data automatically and create more labelled examples to augment the original dataset. This application is especially useful in domains such as medicine where labelling a large set of data could be time-consuming and expensive \citep{frid2018synthetic, sandfort2019data, sela2017gazegan}.
Moreover, research in the same domain has shown that the generated datasets could even be used directly without the original dataset for privacy protection. For example, accurate but fake medical images can be generated instead of using actual images to protect the identity of patients, and it has been proven that the training accuracy of using generated labels in lieu of real images could approach the accuracy of training with real data while protecting sensitive privacy data of patients at the same time \citep{beaulieu2019privacy, bowles2018gansfer}.

Inspired by the extended applications of GANs from other fields, we seek to investigate the scientific applications of GANs in GIS. Instead of transferring stylistic representations between maps with the same level of information, we investigate whether GANs can be used to generate realistic geospatial features by translating a more readily available dataset to one that is available less frequently. We believe that this idea could potentially introduce a new generative cartographic method and unpack many other applications such as contributing to the completeness of data or upsampling existing data sources for downstream geospatial deep learning tasks.

Our motivation, geared towards enhancing the scale, completeness, and array of existing data, is driven by recent rapid developments of acquiring certain types of geospatial data and at a coarse resolution, reaching near-global coverage and completeness.
For example, land use datasets \citep{Robinson_2021,Ludwig_2021}, aggregated urban form \citep{Esch:2017,Chen.2020gnca}, street networks \citep{Zhang_2015,Brovelli_2016} and population data at the district level \citep{Tatem.2017,Lloyd:2019gc} are now available widely and freely for nearly any location worldwide, with recent developments showing no sign of waning.
However, other types of spatial data, such as building footprints, 3D city models and points-of-interest (POIs), which are perhaps more complex and laborious to acquire but no less important for spatial data infrastructures, remain scarce \citep{Bright.2018,biljecki2021open}.
OpenStreetMap (OSM), the principal instance of Volunteered Geographic Information (VGI)~\citep{Yan:2020dp}, is an archetypal example of such heterogeneity: data on roads reached near full completeness \citep{barrington2017world,Minaei_2020}, giving rise to an increasing number of large-scale analyses \citep{Boeing.2021,Calafiore_2021}.
However, data on buildings in the same repository remain missing or partially complete in large swaths of land around the world, even in areas where other features such as street networks are fully mapped \citep{Anderson.2019,barrington2017world,Biljecki.2020,So.2020,Herfort.2021}.

Given the intertwined nature of different features in the built environment across multiple scales \citep{Milojevic-Dupont.20200hc,Majic.2021,Mocnik.2021}, we put forward a hypothesis that we can use GAN to translate one spatial dataset from another.
That is, we seek to apply this technique on widely available geographic datasets at the same or coarser scale (i.e.\ street networks and land use) to produce counterparts at a finer scale that are less frequently available (i.e.\ footprints of individual buildings).

In this paper, we focus on buildings because they are one of the most prominent features in the built environment and data on buildings is being relied upon a rapidly growing number and variety of applications \citep{Fleischmann:2020fe,Patias.2021,Palliwal.2021,Botta.2021,wu2020roofpedia,Harig_2021}, but remains constrained by the current landscape of data worldwide, which is lagging in completeness and coverage in comparison to other features.

As we develop a novel approach to map individual building footprints at a large scale, we also endeavour to design and implement a scalable and generalisable approach that will be replicable with other types of features, therefore seeking to contribute to the field with a new method of mapping spatial data through translation with GANs.

While novel AI techniques have been extensively applied in collecting data on buildings (e.g.\ footprints and heights), nearly all approaches require other forms of comparably scarce or complex data such as point clouds and high-resolution imagery, or they operate at a coarse scale such as deriving the rough urban fabric without delineating individual buildings.

Thus, they perform either at high resolution but limited scale or at large-scale but at low resolution \citep{Geis:2019,Bshouty:2019ei,Park:2019cf,Xie.2019,Sun.2020,Li:2020,Dollner:2020is,Chen.2021r5p,Fan.2021,Ledoux.2021}, leaving no means to generate higher resolution data on buildings at large-scale in an inexpensive and feasible manner.
Those approaches that are demonstrated to perform at higher resolution (regarding individual buildings) and have a potential to scale widely, rather focus on enriching existing data on buildings (e.g.\ predicting additional attributes) instead of generating entirely new data \citep{Milojevic-Dupont.20200hc}. Our approach instead seeks to use widely and freely available data such as land use and road data to create maps of buildings from scratch.

\begin{figure}[h]
\centering\includegraphics[width=0.8\linewidth]{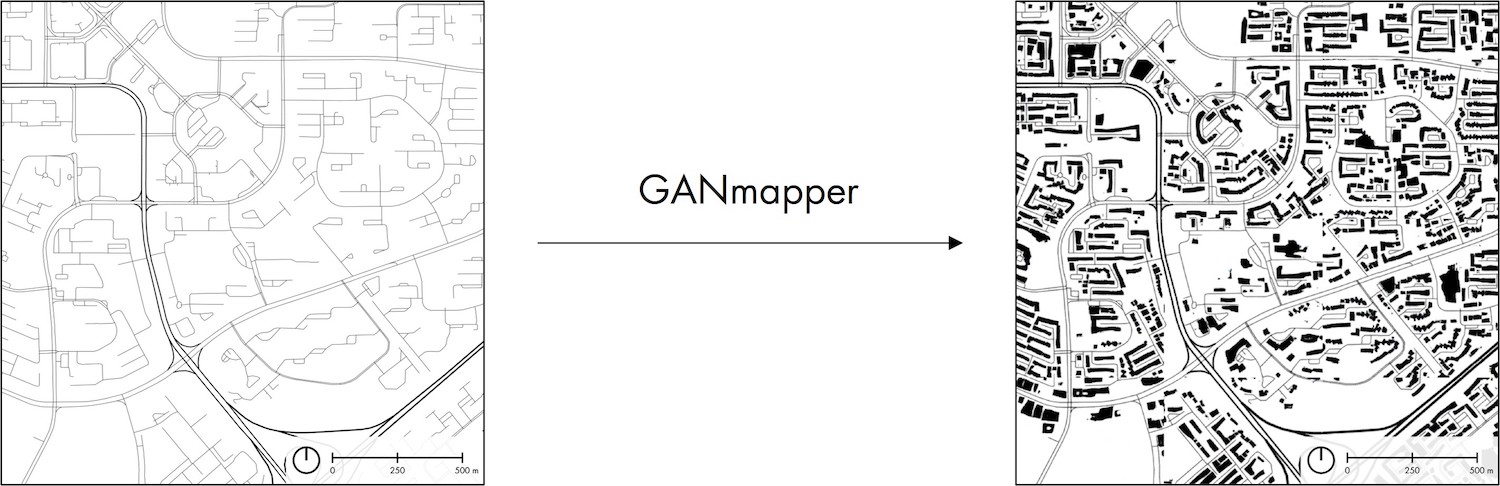}
\caption{Our work in a nutshell: we investigate the use of GAN in generating a spatial dataset from another one. Such an approach may enable translating widely available data such as road networks to create maps of less common features such as building footprints.}
\label{fig:summary}
\end{figure}

In this paper, we introduce GANmapper, a generator of spatial data on buildings (i.e.\ footprints) that is capable of creating realistic looking and hierarchically correct building footprint data from other related urban datasets such as street networks (Figure \ref{fig:summary}).

The paper is organised as follows.
Section~\ref{sc:background} provides an introduction to GAN and uses hitherto in the domain of spatial sciences documented in international scientific outlets.
Section~\ref{sc:methodology} presents the data preparation technique and our proposed model architecture.
Section~\ref{sc:results} discusses the results and demonstrates the scalability and fidelity of using GANs in generating missing spatial building information in the urban context, together with a discussion on potential applications.
We conclude the paper with Section~\ref{sc:conclusion}.

\section{Background}\label{sc:background}
\subsection{Generative Adversarial Networks}\label{sc:gan-intro}

Traditionally, generative models in the field of machine learning are unsupervised where the model would automatically learn the regularities or patterns in the input data in such a way that it can be used to generate new examples that closely represent the input data \citep{han2018unsupervised, pmlr, maaloe2016auxiliary}. 

GANs, on the other hand, added a supervised task to the generative process by using two models in the training process. A generator and discriminator pair based on game theory is used where the generator network must compete against the adversarial discriminator \citep{goodfellow2014generative, goodfellow2016}. The generator network directly produces samples. Its adversary, the discriminator network, attempts to distinguish between samples drawn from the training data and samples drawn from the generator. 
Enhancing the generator-discriminator architecture with deep convolutional neural networks \citep{radford2015unsupervised}, state-of-the-art unconditional GAN models could generate high-fidelity photo-realistic images that could pass the scrutiny of untrained eyes as real images \citep{brock2018large, karras2019style, karras2020analyzing, zakharov2019few}.

While unconditional GANs generate images out of randomly initialised noises, conditional GANs allow granular semantic control of the generated images \citep{pix2pix2017, park2019semantic}. During training, a mask-image pair is used for the generator model to learn the weights to translate the input images to realistic-looking output images. This approach allows additional layers of control and users would be able to specify different source-target combinations for a variety of use cases. 

Both conditional and unconditional GANs have shown promising applications in a multitude of fields. 
Apart from generating new data, as hinted at in the introduction, GANs can also outperform traditional techniques in content translation or content filling. For example, cycle-consistent adversarial network (CycleGAN), proposed by \citet{zhu2017unpaired}, is able to translate photos into different stylistic expressions, and text-based GANs are able translate text inputs into images \citep{zhang2017stackgan, reed2016generative}.
Furthermore, using a tweaked architecture, GANs can be used in content-aware image inpainting to fill up missing areas in an image with high fidelity \citep{li2017generative, yeh2017semantic, pathak2016context}.

Moreover, GANs are not restricted to generating image data. With an appropriate generator architecture, GANs can be used to generate sequential data such as audio and music as well. They are now able to generate melodies from lyrics \citep{yu2021conditional} or compose music from a single latent vector \citep{engel2019gansynth}, or other synthetic time-series data like electrocardiograms \citep{zhu2019electrocardiogram}, stock market trends, and electricity consumption \citep{yoon2019time}.

\subsection{Applications of Generative Adversarial Networks in GIS}\label{sc:works}

Researchers have recently begun to investigate the applications of GANs to geospatial data.
For example, unconditional GANs have been used to generate realistic synthetic satellite images of landscapes and cities from randomly initiated noises \citep{abady2020, zhao2021}, and conditional GANs have been explored to translate satellite images into cartographic representations \citep{pix2pix2017, li2020mapgan} or generate cartographic representations from geospatial vector data \citep{Kang.2019}. Conditional GANs can also create semantic-responsive land-cover maps with user-drawn colour masks, adding human input in the generative process \citep{park2019semantic, Baier_2021}. \citet{9205243} also explored translating historical maps to satellite images to raise the awareness of shifting landscapes. Furthermore, using CycleGAN \citep{zhu2017unpaired}, \citet{Ye.2021} have created an automatic colouring GAN that creates colour renders of urban masterplans from CAD drawings instantly.

While the above-mentioned models have gathered wide interest, their aesthetic value and novelty often exceed their practical use, and the scientific aspect of GANs in GIScience and cartography still remains underexplored. Inspired by many convincing use cases of GANs in other fields, we believe that GANs can be used to create or augment geospatial data and maps through translating one dataset to another. Uncovering such applications may open up many opportunities in downstream research. 

There are currently a few notable papers that explore generating or interpolating spatial datasets with GANs. \citet{Courtial.2021} generated topographic maps of urban areas that follow topographical constraints and preserve the structure, orientation and relative density of the training samples. This approach allows researchers to `summarise' maps with detailed building footprints into more general urban forms for ease of interpretation. Using a similar approach, \citet{Zhu.2019} demonstrated the ability of GANs in interpolating elevation data and achieved good results compared to statistical methods. 

There are also examples where GANs are used to enhance satellite imagery in remote sensing tasks. \citet{9554491} showed that GANs can be used to restore deformed satellite images caused by jitters in geopositioning. Apart from distortion restoration, GANs can also outperform state-of-the-art methods in satellite image sharpening and superresolution \citep{MA2020110, 8677274}. In addition, GANs can also be used in image data cleaning. For example, \citet{LI2020373} used a GAN to remove cloud contamination in remote sensing images and \citet{9517298} has developed a method that could automatically remove unwanted objects in street view imagery \citep{Biljecki.2021}.

Besides applying GANs in cartography and remote sensing, the research of GANs on spatiotemporal data has also been on the rise. \citet{Rao2020} used a GAN with a Long Short-Term Memory (LSTM) Architecture \citep{hochreiter1997long} to generate synthetic commuter trajectory data to facilitate downstream learning while preserving user identity, which is relevant in the current day of ever-increasing concern on user data privacy, especially if including location data. With a similar approach, \citet{9216472}, \citet{XU2020102635}, and \citet{LSTM1} used spatiotemporal GANs to generate or predict traffic flow data for virtual training and transport system simulation which could help the design of a city's transportation system.

\section{Methodology}\label{sc:methodology}

\subsection{Overview}

Inspired by the ability of GANs to generate reliable synthetic data in other fields, we seek to explore the feasibility of using GANs to generate geospatial data. We argue that the application of GANs in GIScience and cartography should not remain in merely transferring one mapping style to another (Section~\ref{sc:works}) but can transcend on to be used to generate spatially correct synthetic datasets by taking cues from another spatial dataset. 

We call the process of synthesizing one geospatial dataset from another related geospatial dataset `geographical data translation', and we propose GANmapper, a geographical content translation AI based on a GAN architecture that is able to translate street network data into synthetic building footprint data that is visually and morphologically similar to the ground truth.

To investigate the performance and scalability of our model, we create four different experiments with OSM data from nine cities across the world representing different forms of urban morphology. These cities are major cities in the world that generally have a high degree of building data completion to provide adequate training data and ground truths for evaluation.

By focusing on OSM data, our work essentially investigates the potential of using one set of (better mapped) features such as street networks to fill another (less or entirely unmapped) counterpart, levelling their completeness and quality.

Building footprints in datasets such as OSM may be entirely missing or have partial completeness in some locations~\citep{Fan:2014kz,biljecki2021open}.
Our approach is designed to work in unmapped areas, ignoring existing footprints, if any.
Thus, it is important to underline that no building footprints are used as input, presenting an approach to derive new spatial data from scratch.
However, the method would work also in locations with partial completeness of data on buildings.

\subsection{Model Architecture}

\begin{figure}[h]
\centering\includegraphics[width=1\linewidth]{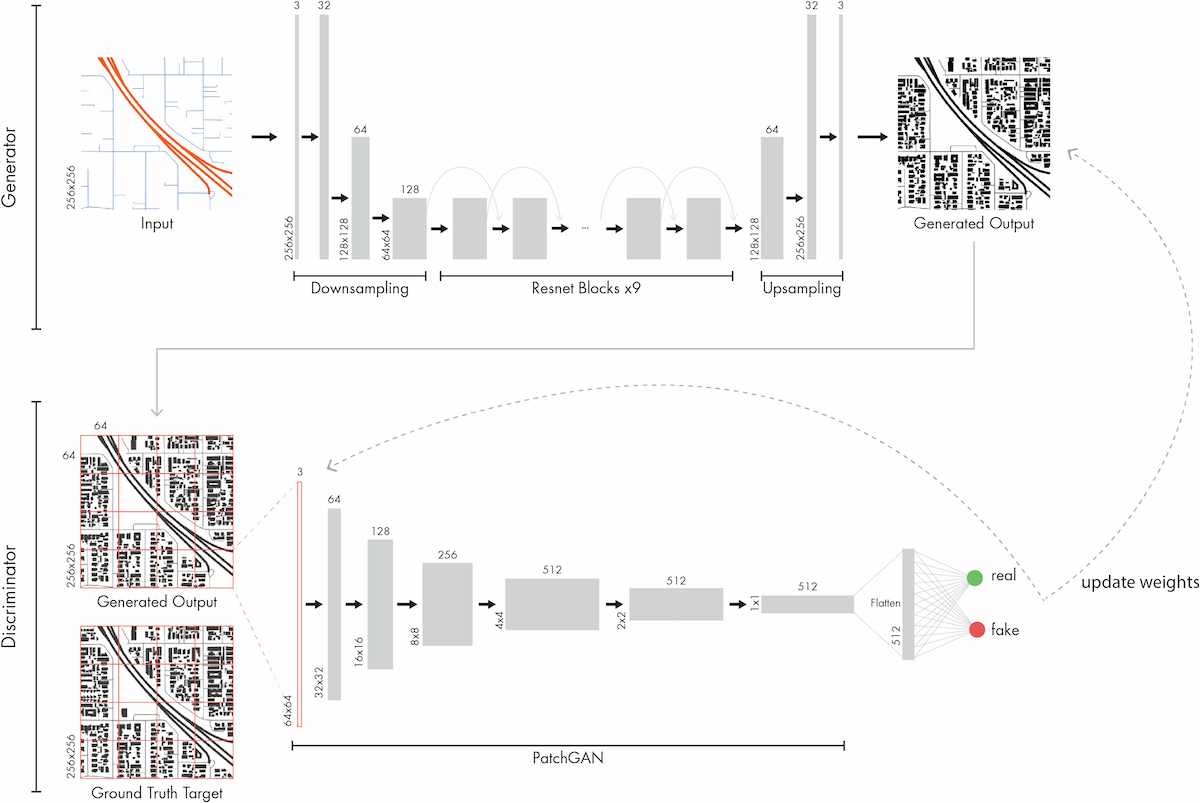}
\caption{The GANmapper model architecture.}
\label{fig:architecture}
\end{figure}

The GANmapper model architecture is a type of Image-to-Image Conditional GAN \citep{mirza2014conditional, pix2pix2017} that translates input image data such as street networks to a target image populated with generated building footprints. According to our literature review of the international scientific literature (Section~\ref{sc:works}), this work is the first one investigating the performance and possibility of such a use case with conditional GANs. 

Figure \ref{fig:architecture} provides an overview of the proposed GAN architecture. We use an autoencoder \citep{hinton2006reducing} architecture with 9 residual blocks \citep{he2016deep} for the generator.
In each forward pass, the generator will try to generate outputs that could `fool' the discriminator into classifying the generated images as `real', while the discriminator will learn to classify the generated images as `fake' and ground-truth targets as `real'. The discriminator splits the input image into 70x70 tiles and outputs a Boolean classification for the generated image from the average classification value of all split tiles. At the end of each forward pass, the losses for both the generator and discriminator is calculated and their weights updated. As each epoch passes, both the generator and discriminator get better in their roles until the generator's output would pass the scrutiny of the discriminator as a real image 50\% of the time as in a zero-sum game.

\subsection{Pre- and post-processing}\label{sc:process}

\begin{figure}[h]
\centering\includegraphics[width=1\linewidth]{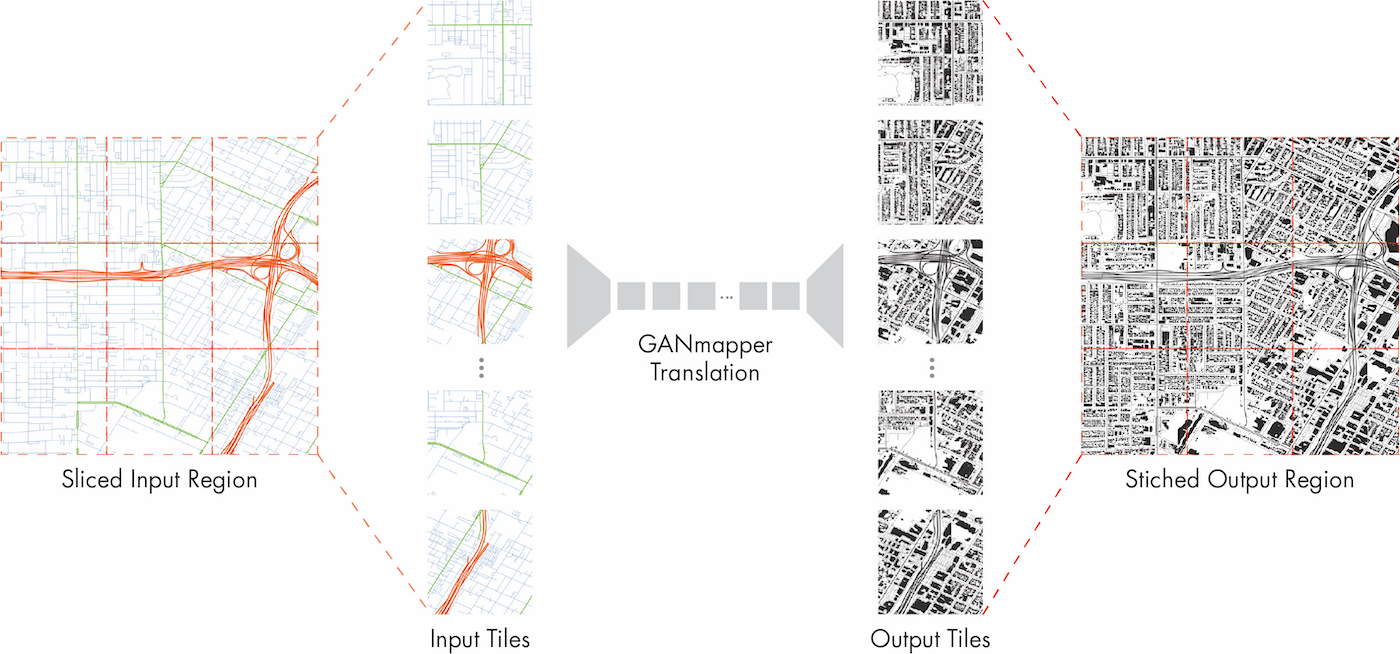}
\caption{Pre-processing and post-processing for large regions.}
\label{fig:pipeline}
\end{figure}

As with other deep learning models, the resolution of the input and output is limited by the memory of the hardware, and the resource needed scales proportionally to the size of the input images. Thus, it would be too resource-intensive to fit a large area (e.g.\ an entire city or a neighbourhood) within a single high-definition image. 

The same problem of large sized images for raster maps was also encountered in the earlier days of web map services. Web map providers such as Microsoft and OSM have since adopted the Web Map Tile standard which allowed large maps to be broken down into smaller tiles for faster loading \citep{Garcia12, tilemaps}. Some previous works in large scale object detection ins satellite images also utilised this processing method in training their deep learning models~\citep{Hofmann, wu2020roofpedia}.

Inspired by these efforts, we created a pre-processing pipeline that converts the targeted areas into 256x256 raster tiles into a XYZ tile directory, which is one of the standard formats of the Web Map Tile standard. The output of the models are also generated according to the same directory structure as the input tiles. Therefore, the output images can easily be restored into a larger whole for easy visualisation and portability (Figure~\ref{fig:pipeline}). 

\subsection{Datasets and Preliminary Exploration}\label{sc:data}

\begin{figure}[h]
\centering\includegraphics[width=\linewidth]{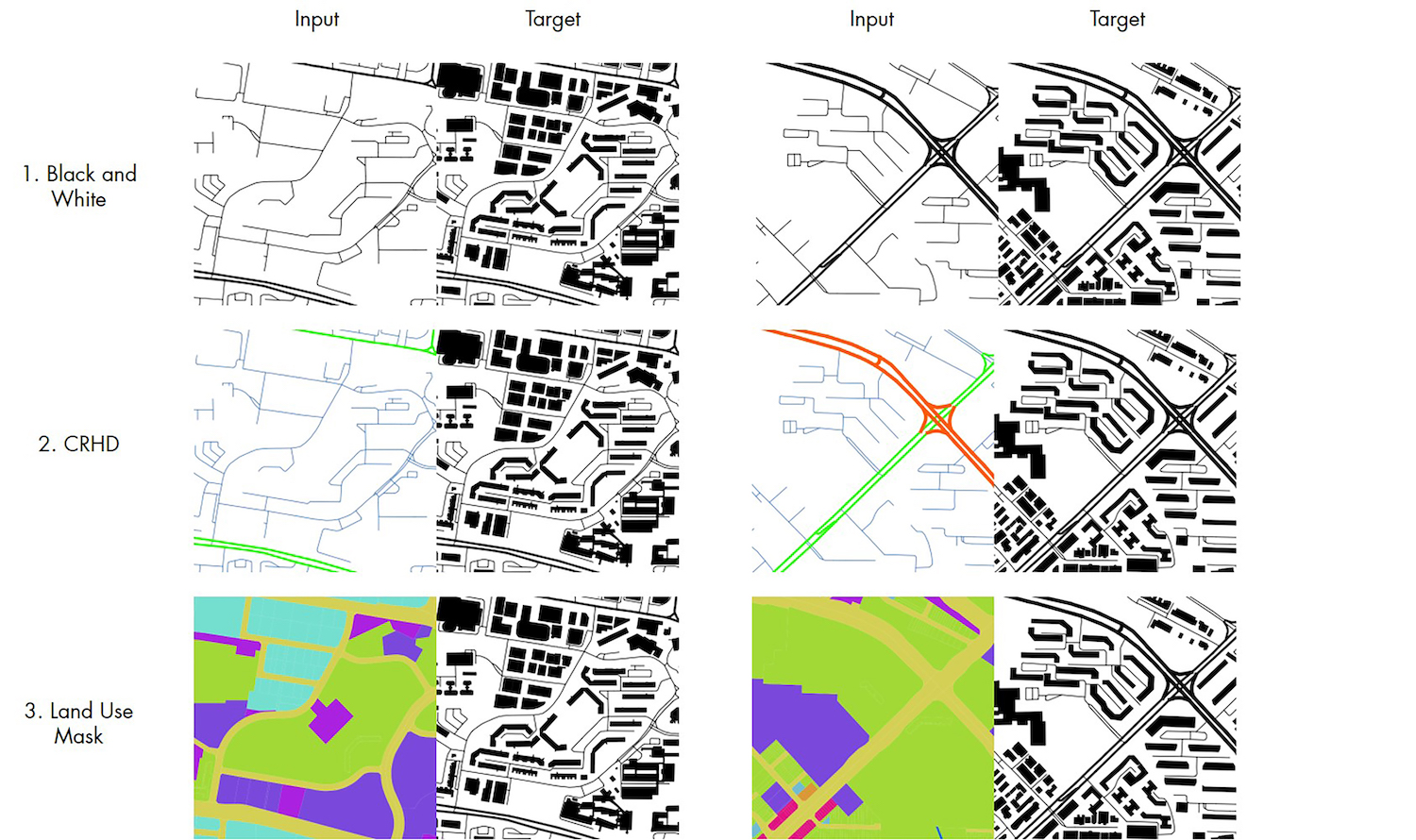}
\caption{Three different input types, which we investigate independently of each other. While each input dataset provides promising results, in the experiments, the Coloured Road Diagrams (CRHD; in the second row) proved to be the best performing input dataset. Thus, we focus on using this type of representation in the detailed experiments.}
\label{fig:data}
\end{figure}

Urban morphology can be encapsulated with a variety of cartographic representations. In the model, the input data provides the initial conditions to the generator and affects the ability of the model to learn efficiently.

As this paper is perhaps one of the first few attempts in geospatial data translation with GAN, especially with image data, we seek to investigate the most suitable input-output pair for the task. We experimented with three different input representations derived from two datasets that are widely available: street networks (two variants) and land use data.
In these preliminary explorations, we focus on Singapore.

Figure~\ref{fig:data} shows the different input-target pairs used in the experiments for training. OSM road network data (first two rows) are used to create two types of inputs: (1) Single-Channel Black and White representation with road hierarchy differentiated by line-width; and (2) Coloured Road Hierarchy Diagrams (CRHDs) that represent road hierarchy in colour on top of line width~\citep{chen2021classification}.
The target images are composed of Black and White input images with building polygons obtained from OSM.

The land use map of Singapore is used to create a third input-target pair as an alternative representation.
For this purpose, we have used open data\footnote{\url{https://www.ura.gov.sg/Corporate/Planning/Master-Plan}; last accessed on 28 July 2021.} by the Urban Redevelopment Authority of Singapore.
The different land use parcels are colour-coded and represent different land use schemes. This image mimics the image-mask pair used in other conditional GANs such as pix2pix for a comparative study on the effect of input types on the prediction results. 

\subsection{Experiment Setup}

Four experiments are set up with different configurations and datasets to investigate the performance and feasibility of our method.

In Experiment 1, we investigate the performance of the three types of input-output pairs mentioned in the previous section to find the best performing pair. We also compare the results of our model in all three input-output pairs with a popular baseline Conditional GAN model proposed by \citet{pix2pix2017}.

In Experiment 2, we use the best performing input-output pair found in Experiment 1 to compare the performance of the results on 4 different tiles scaling configurations, ranging from a neighbourhood scale of about 2km per tile to the street-level scale of about 300m per tile (zoom levels 14, 15, 16, and 17). 
The results of this experiment will help us understand how the method is influenced by the different zoom levels of the data (i.e.\ scale or level of detail), and whether the method is more reliable at generating small-scale maps or it can be used for large-scale maps in which the level of detail of building footprints is finer.

With an understanding of the performance of the model across different zoom levels, Experiment 3 applies GANmapper to 9 different cities to investigate its scalability and versatility.
These 9 cities are from around the world, representing different urban forms: Beijing, Jakarta, Los Angeles, London, Munich, New York, Paris, Singapore, and Tokyo.
The results will be compared across cities to assess the performance and robustness of our model in different morphological conditions around the world.

Finally, Experiment 4 investigates the quality of the generated data when the constituent tiles are stitched back together by the post-processing pipeline. We chose one region in each of the 4 different cities (Singapore, Jakarta, Paris, Los Angeles) to assess the spatial consistency of the assembled regions.

In all the four experiments, the model is trained with a train-test split of 80\%/20\% and the size of the dataset is recorded in each subsection.

\subsection{Metrics for evaluation}

Frechet Inception Distance (FID) proposed by \citet{heusel2017gans} is one of the most commonly adapted benchmarks for GAN performance. FID assesses the quality of images created by a generative model by calculating the distance between the feature vectors of real and generated images that are computed using an Inception-v3 \citep{szegedy2016rethinking} image classification model.
A lower FID score indicates that the two groups of images are more similar in terms of the extracted feature vectors with a perfect score of 0.0, indicating that the two groups are identical.
Visually, a lower FID score on the generated images tends to correlate well with realistic-looking images, indicating more morphological correct representations in terms of the general shape, size, and density of the generated building footprints.
Mathematically, FID can be expressed with the following formula:

\[FID = \left \| \mu_r - \mu_g \right \| ^2 + Tr(\sum r + \sum g - 2\sqrt{\sum r \sum g})\]
where $X_r\sim N(\mu_r,\sum r)$ and  $X_g \sim N(\mu_g,\sum g)$ are the 2048-dimensional activations of the Inception-v3 pool3 layer for real and generated samples respectively.

In addition to FID, the Mean Intersection over Union (mIoU) between the inputs and generated images is also calculated in some experiments to measure the degree of overlap of the generated images to the ground truth. Mathematically, mIoU measures the number of pixels common between the input and generated images divided by the total number of pixels present across both datasets. 

A high mIoU score indicates that the generated images overlaps with the input more and thus replicates the input data better. However, due to the generative nature of GANs, the generated results should look realistic in terms of the overall building morphology but not overlap exactly with the input image. If the generated image has a high mIoU score (close to 1), we can conclude that the generator is overfitted to the training data, which impedes the models' ability to generalise its learned patterns to new inputs.

\subsection{Implementation}

The model is implemented with PyTorch \citep{NEURIPS2019_9015}, and QGIS is used in data processing. We release GANmapper as open-source software.
To maintain the consistency of all experiments, we trained the models at the same resolutions with the same parameters at 100 epochs regardless of the dataset. Images shown in the experiments are picked from test sets which area isolated from the training set. 
Further, in the repository, we also include the pre-processing and post-processing pipelines, which may aid other researchers to extend the work.

\section{Results and evaluation}\label{sc:results}
\subsection{Experiment 1 --- Finding the best input configuration}\label{sc:exp1}

In Experiment 1, we explore two different types of inputs: street networks and land use parcels to assess their suitability for building footprint generation.

As mentioned in Section~\ref{sc:data}, the street network inputs are generated from different types of roads in a city. The primary roads are usually highways or expressways that connect different districts; the secondary roads are usually avenues that connect neighbourhoods and often demarcate the boundaries of one neighbourhood to another; the tertiary roads are capillaries that penetrate the neighbourhoods. Such street network maps are often used to classify urban morphology and inform the urban granularity and density of an area. 

On the other hand, the land use map of Singapore used in this experiment only offers road network information until the secondary level. Tertiary capillaries are replaced by colour-coded land use parcels that demarcate the building type and their gross floor area ratio (GFA). These two factors play a significant role in affecting the eventual building footprint in the parcel. For example, a large parcel of land for residential use with a low GFA limit will often result in small, scattered houses, whereas a smaller parcel of land with the same GFA limit could become a high-rise development.

\begin{figure}[h]
\centering\includegraphics[width=\linewidth]{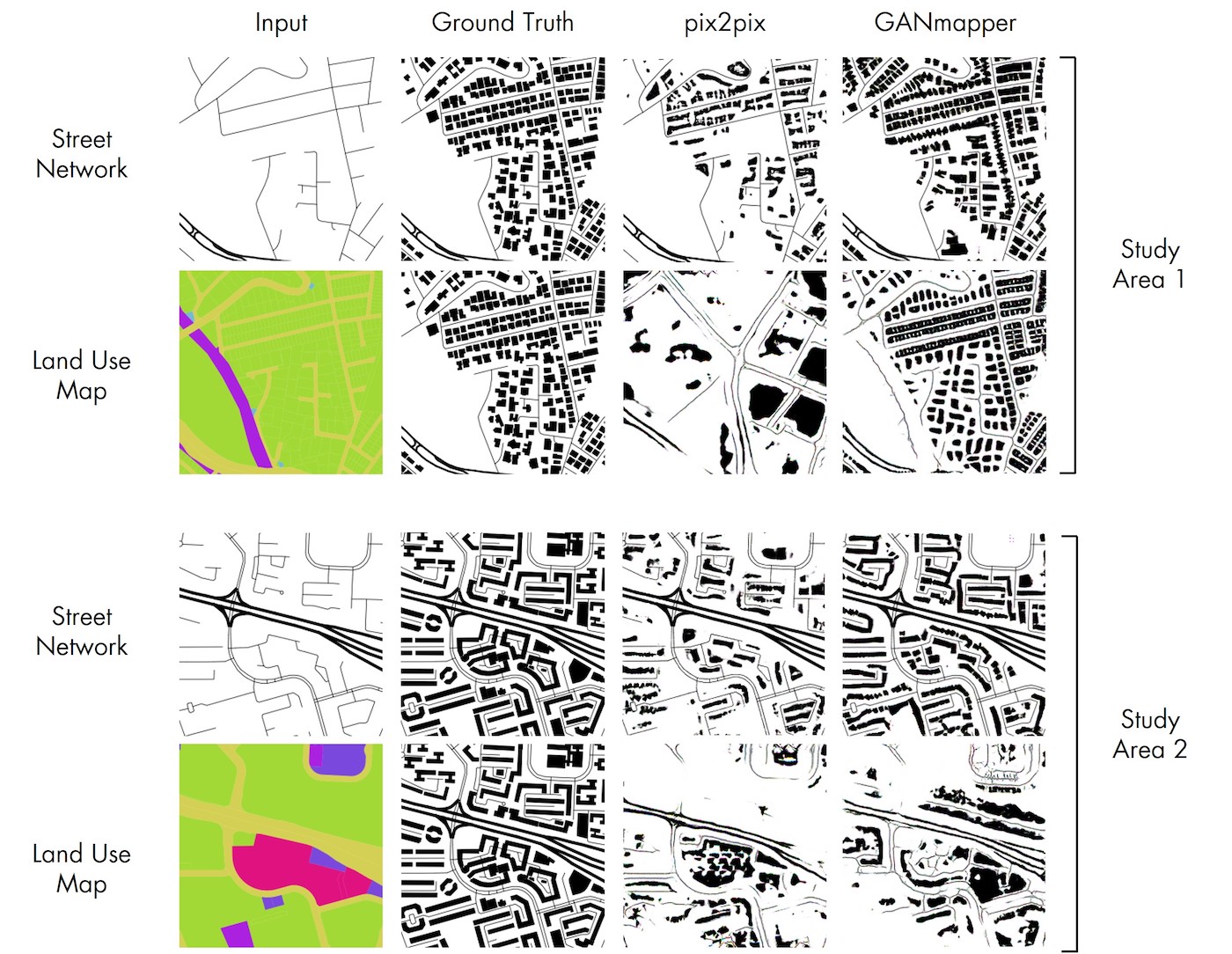}
\caption{Results of using different input types: street network vs landuse maps, together with a comparison with a baseline Conditional GAN (pix2pix), which our custom model outperforms. The input (open) data in this example is courtesy of OpenStreetMap and the Urban Redevelopment Authority of Singapore.}
\label{fig:2input}
\end{figure}

Figure \ref{fig:2input} shows a sample of the results from the two types of inputs. Results from the original pix2pix model architecture \citep{pix2pix2017} using a U-net \citep{ronneberger2015u} generator architecture are also included as a baseline. From the figure, we see that the GANmapper architecture generates more realistic-looking images with sharper building footprint edges and better reproduction of the ground truth's urban texture compared to the pix2pix baseline, affirming the importance of developing an approach tailored for geographic information. 

In both models, the street network input returns better visual performance compared to the land use map input. This difference is likely due to the fact that the model is not yet able to extract representative features for each type of land use parcel. It is also possible that the colour mappings do not provide sufficient information to the model on the building morphology. For example, the green parcels representing residential areas could be both landed houses or elongated apartment blocks, causing confusion in the case of the generated image from the land use map in Study Area 2.

\begin{figure}[h]
\centering\includegraphics[width=0.9\linewidth]{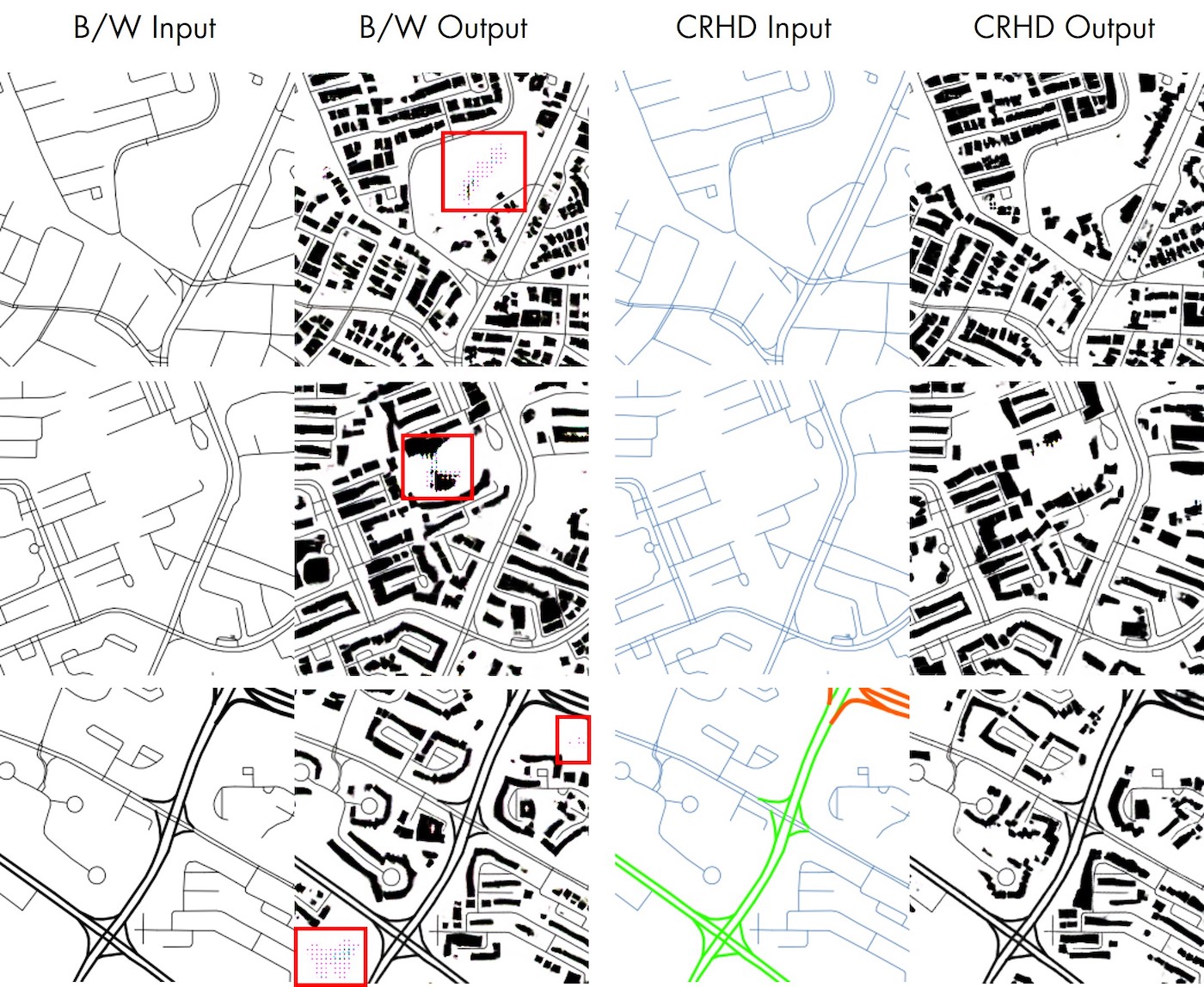}
\caption{CRHD inputs minimise checkerboard artifacts highlighted in red boxes, which are typical in GAN outputs.}
\label{fig:CRHD}
\end{figure}

However, the Black-and-White street network is not without flaws either. Figure~\ref{fig:CRHD} (on the left) shows the presence of artifacts in a sample of generated images from the Black-and-White (B/W) input. The appearance of artifacts such as the ones highlighted in the figure is generally due to the over-fitting of the generator and confusion of the discriminator having too little information to differentiate the real from the fake. 

This problem can be ameliorated by using Coloured Road Hierarchy Diagrams (CRHD) proposed by \citet{chen2021classification}. CRHDs adds colour to differentiate hierarchies in the road network on top of using road thickness. The addition of colour creates 3-channel images which provide additional information to the input to avoid overfitting of the generator under the same training settings. As seen in the Figure~\ref{fig:CRHD}, outputs using CRHDs do not experience artifacts in vacant areas. 
Therefore, in the continuation of the work, we opt for images of street networks that signify the types of roads using colours.
Such images are fairly easy to generate and the data on the road type is widely available, providing a substantial advantage over their B/W counterparts.

\begin{table}[h]
  \centering
\footnotesize
  \caption{FID Scores based on input types.}
    \begin{tabular}{lrrrr}
    \toprule
    \textbf{Input Style}  & \multicolumn{2}{c}{\textbf{FID}} & \multicolumn{2}{c}{\textbf{mIOU}} \\
    \cmidrule(lr){2-3}\cmidrule(lr){4-5}
    \textbf{ }  & \textbf{pix2pix} & \textbf{Ours} & \textbf{pix2pix} & \textbf{Ours} \\
    \midrule
    Roads (B/W)   & 131.8 & 104.4 &  0.565 & 0.420 \\
    Roads (CRHD)    & 134.0 & 98.3  &  0.455 & 0.484 \\
    Land use & 216.2 & 243.4 & 0.497 & 0.457 \\
    \bottomrule
    \end{tabular}
  \label{tab:exp1}
\end{table}

Comparing the average performance of the three types of inputs over all images in the test set with FID and mIoU score (Table~\ref{tab:exp1}), CRHD input style returns the lowest FID score for our model architecture. 
A lower FID score indicates that the extracted feature vectors of the generated images match the extracted feature of the ground-truth images more closely, resulting in more realistic and ground-truth look-a-like images. This is resonant with the higher visual realism compared to land use maps in Figure~\ref{fig:2input} and the removal of checker-board artifacts in Figure~\ref{fig:CRHD}. 
Looking at mIoU score that measures how close the generated pixels overlap the ground truth, pix2pix with B/W street input performs the best with a score of 0.565, while the highest score obtained by our model is 0.484 with CRHD input. The low value for mIoU score is expected as the loss function of GANmapper measures the real/fake in terms of visual similarity rather than pixel-wise similarity. This difference means that although the generated image could have a high FID score with stylistically similar urban grain and density to the ground truth, the generated footprints might not overlap the ground truths footprint closely.

Since FID scores appraises images based on visual affinity, it can be said that it is a good measure of the morphological similarity between the generated dataset and the ground truth. The differences in the locations of individual polygons should not have a strong impact on the proposed uses of GANmapper at the urban level. Moving forward, we will focus on using FID scores as the metric to measure visual and morphological similarity.

\subsection{Experiment 2 --- Understanding the effect of tile scaling}\label{sc:exp2}

As elaborated in Section~\ref{sc:process}, large areas need to be sliced into tiles for prediction. Without expanding or contracting a study area, the scale of the tiles or `zoom levels' according to slippymap's convention would affect the resultant size of the dataset. For example, a fixed area of 1 kilometre square will produce 4 images (each tile will have a width of about 500m) and 16 images at a finer scale (around 250m per tile). 

Another factor that needs to be considered is the resolution of the input images. At 256x256 px resolution, fitting too large an area could result in loss of information. Therefore, we need to experiment the ideal combination of the spatial extent and consequently, the level of detail. 

\begin{figure}[h]
\centering\includegraphics[width=1\linewidth]{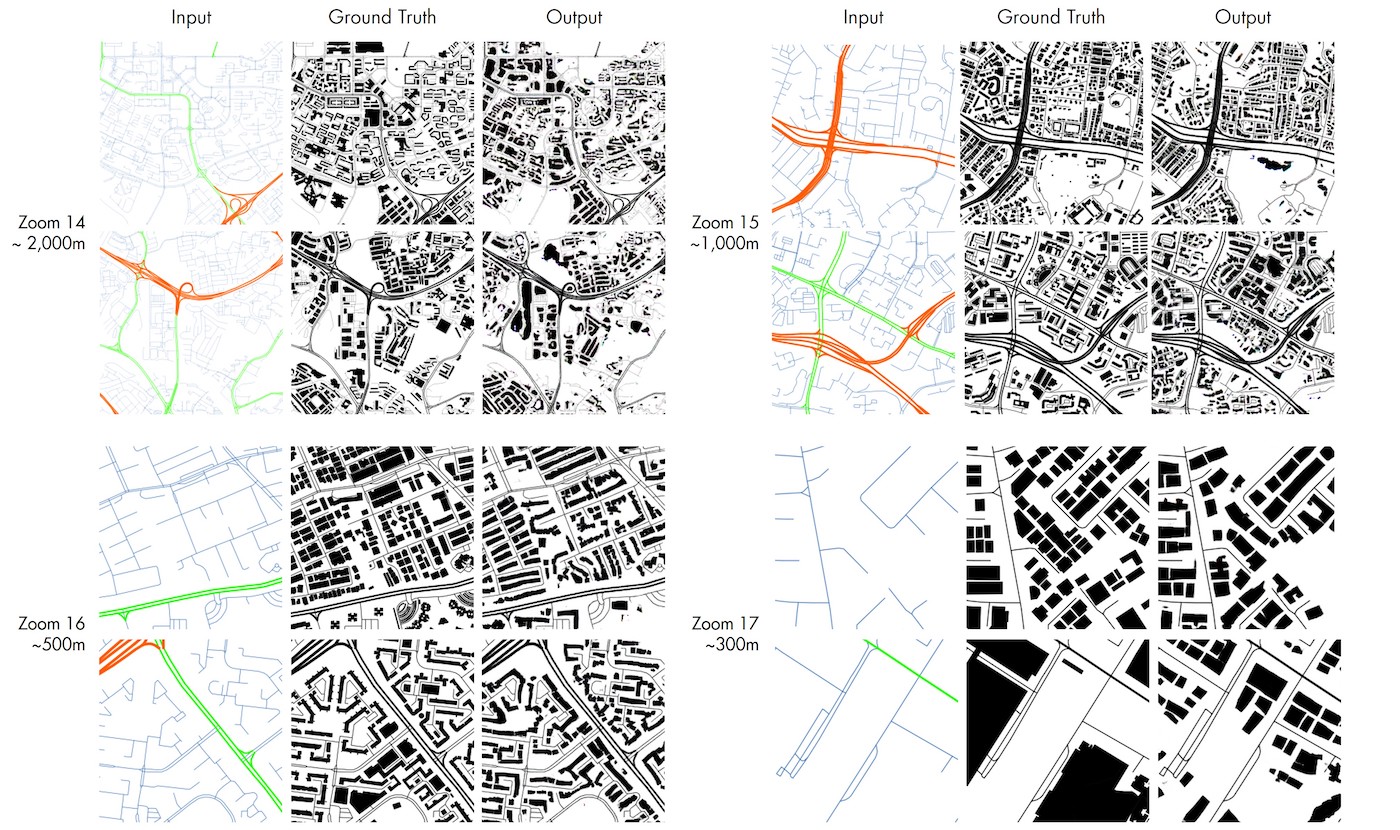}
\caption{Four different zoom levels according to tiling conventions are explored in the experiment. The values in metres indicate the width/height of each tile, suggesting the scale of each tile at its specific zoom level.}
\label{fig:zooms}
\end{figure}

To investigate the performance of the methodology under different scales, we conduct training on 4 different datasets at the slippymap zoom level of 14, 15, 16, and 17 with CRHDs. Table~\ref{tab:exp2} tabulates the scores and training sizes of the different zoom levels in the same area and Figure~\ref{fig:zooms} shows the different scales of the levels and the quality of the generated outputs. 
The same figure also illustrates the size of the spatial extent of each level.

\begin{table}[h]
  \centering
  \footnotesize
  \caption{FID Scores by zoom level.}
    \begin{tabular}{llrrrrr}
    \toprule
    \textbf{Zoom} & \textbf{Scale} & \textbf{Training} & \textbf{B/W} & \textbf{CRHD} & \textbf{B/W} & \textbf{CRHD} \\
    \textbf{ } & \textbf{(tile width)} & \textbf{Size} & \textbf{FID (Test)} & \textbf{FID (Test)} & \textbf{FID (Full)} & \textbf{FID (Full)} \\
    \midrule
    14  & 2,000m & 150  & 123.56 & 137.32 & 72.45 & 80.74 \\
    15  & 1,000m & 394  & 133.01 & 108.26 & 75.56 & 52.39 \\
    16  & 500m & 864  & 83.73  & 77.87  & 42.20 & 38.52 \\
    17  & 300m & 2825 & 63.23  & 52.65  & 38.14 & 24.41 \\
    \bottomrule
    \end{tabular}%
  \label{tab:exp2}%
\end{table}%

Examining both FID scores and visual samples, higher zoom levels (16, 17) performed graphically better than lower zoom levels (14, 15) with sharp footprints and more accurate urban texture. The model trained on zoom level 17 achieved the lowest FID score with morphologically correct and compelling looking building footprints.
Conversely, the model trained on zoom level 14 had the worst FID score and struggles more in producing polygon-like footprints. Nevertheless, it still manages to be morphologically truthful and produces a similar urban texture as the ground truth. 

The reason for the difference in performance is probably due to both the training size and the density of information in a tile. From the entire map of Singapore,  2,825 training images were generated at zoom level 17 while the same region could only produce 150 images at zoom level 14 which is one order of magnitude smaller. 
As with other GANs, a small number of training data could result in overfitting or unstable training, thus leading to worse performance. Furthermore, at lower zoom levels such as 15 and below, the 256x256 resolution of the input-output pair struggles to present the information in high fidelity. This is especially poignant in zoom level 14 where the ground-truth building footprints start to get `blurry' and the hierarchical relationships between roads and building footprints such as the thickness of the road and the street offset demarcated by the blank spaces starts to become too small to be important.

Although zoom 17 performs best in terms of FID score and visual sharpness, it misses out on contextual information such as the second example of zoom level 17 in Figure~\ref{fig:zooms}. In the example, the ground-truth shows a part of a large building footprint while the generated image returned smaller buildings instead as it generates the image solely based on the information of the input but not the surrounding tiles. This lack of contextual information could create inaccuracies in areas with large buildings when the tiles are stitched together which is explored in the next subsection. 

Therefore, the zoom level-accuracy trade-off needs to be considered for each city. In a city with large building footprints, lower zoom levels such as 15 or 16 are preferred to retain contextual information, while cities with smaller grains could still use zoom level 17 for its high visual realism.

\subsection{Experiment 3 --- Applying GANmapper to multiple cities}

So far, the experimental portion of this work focuses on one study area. In this experiment, we test our model's flexibility on a mixture of cities with different urban morphologies and urban textures, to understand how the approach scales elsewhere. Eight additional cities are chosen and each city's data is converted into CRHD datasets with zoom level 16 which offers a balance between training size and quality of prediction according to Experiment 2. Due to variations in urban areas, the size of the datasets varies from city to city. Furthermore, ground-truths that contain no building or a very small number of buildings are removed to avoid learning from incomplete areas. 

\begin{figure}[h!]
\centering\includegraphics[width=\linewidth]{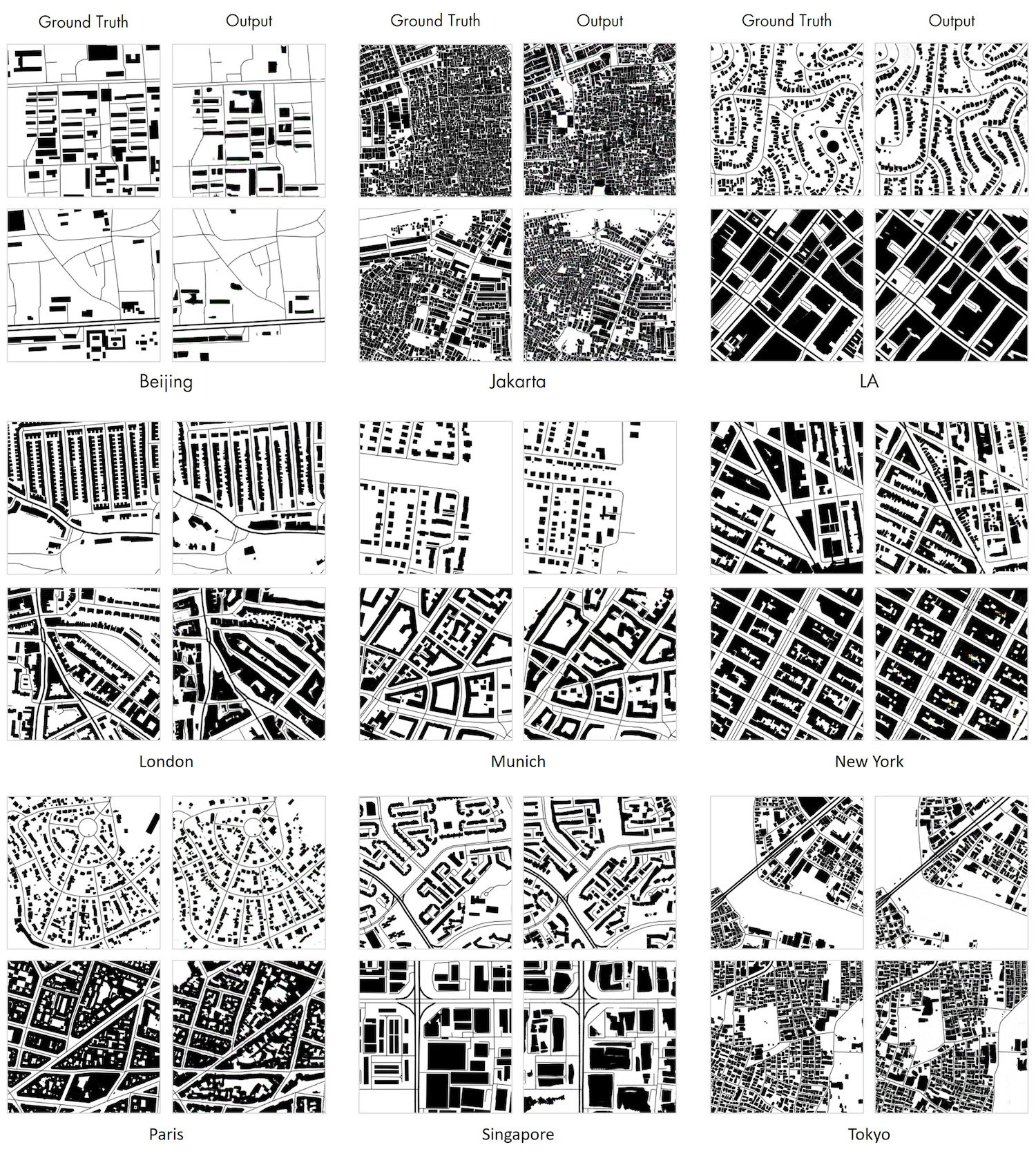}
\caption{Synthetic results of GANmapper, together with the reference data, across several diverse cities around the world at zoom level 16.}
\label{fig:cities}
\end{figure}

\begin{table}[h]
  \centering
  \footnotesize
  \caption{FID scores by city.}
    \begin{tabular}{lrrr}
    \toprule
    \textbf{City} & \textbf{Training Size} & \textbf{FID (Test)}  \\
    \midrule
    Beijing & 832  & 172.28    \\
    Jakarta & 620  & 122.25   \\
    Los Angeles & 1410 & 54.35  \\
    London  & 2147  & 67.15    \\
    Munich & 803  & 68.10   \\
    New York & 1378  &  66.69   \\
    Paris & 2762   & 62.80    \\
    Singapore & 864   & 77.87    \\
    Tokyo & 1323  & 72.81    \\
    \midrule
    \textbf{Average} & \textbf{ } & \textbf{85.11} & \textbf{ } \\
    \bottomrule
    \end{tabular}%
  \label{tab:cities}%
\end{table}%

Table~\ref{tab:cities} records the FID scores of the test sets for the selected cities and Figure~\ref{fig:cities} shows the sample results of the test sets against the ground truths. We observe that the FID score varies across cities. Beijing had the least correct score of 172.3 while Los Angeles had the best score of 54.35.

Visually inspecting the sample results, the model is able to capture the dominant urban textures such as the small-grained settlements in Jakarta, the meandering suburbs of Los Angeles, and the gridiron streets of New York. 
Building footprints, especially those of smaller buildings, are generated as polygons with sharp corners and a certain degree of randomness, which enhances the realism of the generated images. 
The model struggles a little in generating larger footprints. The bigger buildings in Singapore and LA have rounded corners and fuzzier edges that the smaller footprints and the courtyards of New York's townhouses also suffer from fuzziness. 
Nevertheless, in almost all cases, the model respects the roads so there are no overlaps: it is able to learn the appropriate road offsets for each city, and understand the effect of different road hierarchies on the offset and size of the footprints.

The model is also able to infer different densities of the footprints according to the input. In the bottom example of Beijing (top left of Figure~\ref{fig:cities}), the model creates a sparse image even though the road network seems to suggest a neighbourhood-like density. In examples of LA, Paris and Singapore, the model is able to detect subtle morphological differences in the road networks and accurately produce the corresponding building typology and density from the inputs. 

\subsection{Experiment 4 --- Investigating spatial consistency of the stitched outputs}\label{sc:Exp4}

So far, FID scores and visual inspections are conducted on the tile level, and they have demonstrated promising performance. To evaluate the practicality of the model, it is essential to understand its performance when the tiles are stitched together into a larger urban area. In this experiment, we selected 3 disparate models from Experiment 3, Jakarta (FID = 122.25), Los Angeles (FID = 54.35) and Singapore (FID = 77.87) to evaluate the generated outputs against the ground truth after the individual tiles are stitched together. 

\begin{figure}[h!]
\centering\includegraphics[width=0.78\linewidth]{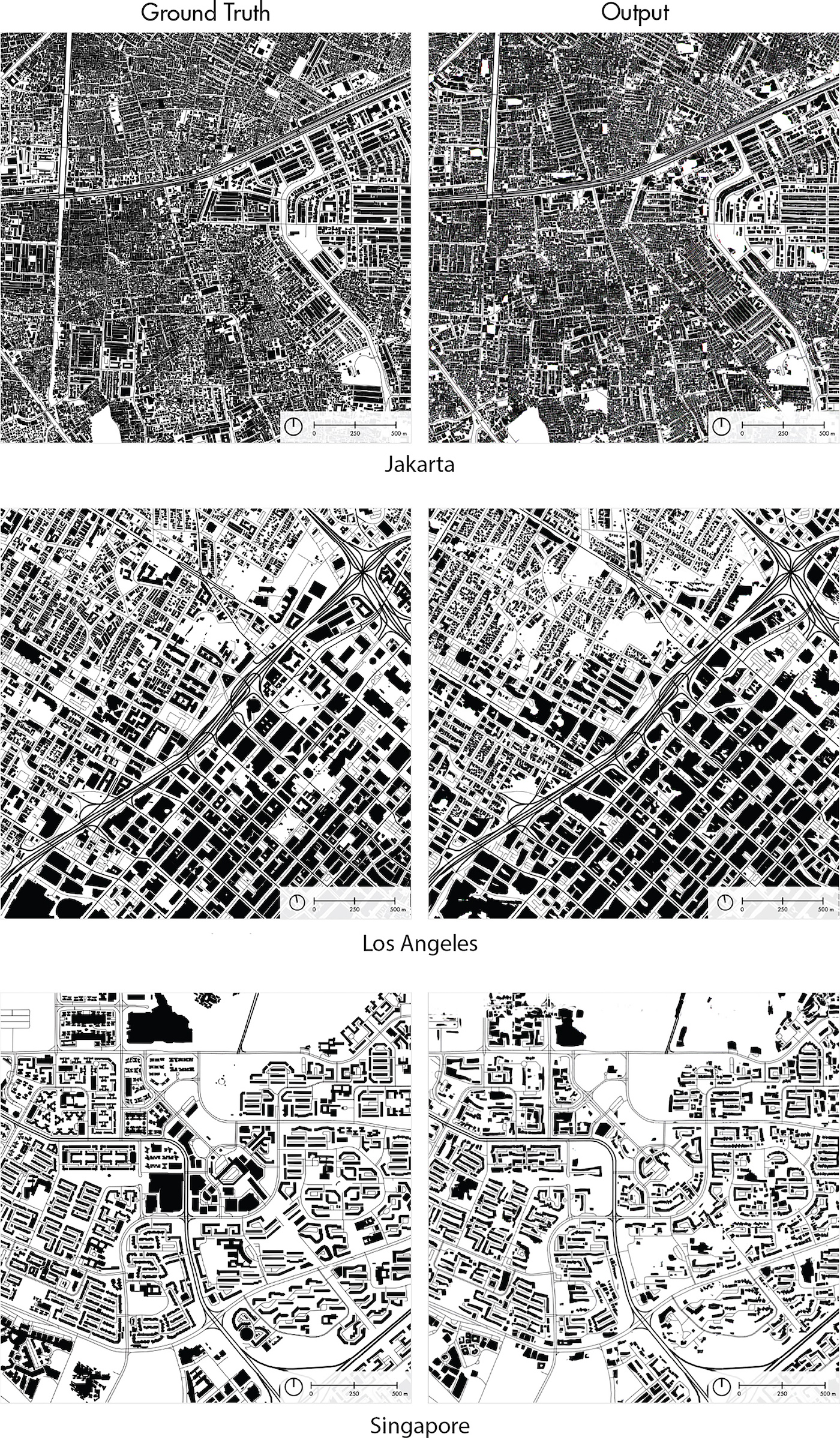}
\caption{The stitched output of Jakarta, Los Angeles, and Singapore.}
\label{fig:exp4all}
\end{figure}

Figure~\ref{fig:exp4all} presents the ground truth-output pairs of Jakarta, Los Angeles and Singapore. All pairs capture an area of approximately 2.5km\textsuperscript{2} and are stitched together from 16 to 25 tiles. Generally, we can observe in the stitched outputs a good reproduction of the correct urban texture in terms of the shape, orientation, density, and area of the building footprints. 

The test area in Jakarta is selected to represent an area with ultra-dense urban sprawl. The structure of the urban form is defined by the primary and secondary roads but the tertiary roads are much more chaotic. We see that the model is able to discern the difference in setbacks between tertiary roads and secondary roads where the settlements are much closer to the tertiary road than those close to the secondary roads. In addition, the model correctly depicts the typology of the area in the east where a curved, river-like void space exists and the blocks generated are much more regular than the rest of the area.  

The test area in LA is chosen to be at an intersection of the urban sprawl and downtown, separated by a highway. We see that the model is able to generate two different typologies on each side of the central highway. The downtown area generally consists of large commercial buildings that occupy an entire street block, whereas the residential area in the north is filled with Single-Family Houses (SFH) typically seen in the LA urban sprawl. 

In Singapore, the test area is a typical public housing town with an amenity node surrounded by long, slab high-rise apartment blocks. The model is able to infer a correct representation of the residential block texture in the area but fails to generate the larger footprints of the amenity buildings at the centre of the neighbourhood. This limitation could be due to the fact that the street morphology near the amenity blocks does not deviate from the rest of the neighbourhood, thus leading the model to generate these high-rise residential buildings rather than shopping malls. On the other hand, looking at the top portion and bottom left of the output where the generated footprints are larger and wider, we can conclude that the model is still able to generate typologies other than residential blocks when the street morphology changes.

\subsection{Discussions and limitations}

From the examples shown in the previous sections, it is clear that our model is able to produce realistic urban texture and building footprints in a variety of urban morphologies and zoom levels. 
The GAN model is able to learn the appropriate sizes and shapes of building footprints and generate clean maps, especially at higher zoom levels. The model is also able to learn the offsets of buildings from the streets and also represent the offsets between the footprints realistically.
When stitched together, the individual images are able to form continuous urban fabric that highly represents the ground truth.
Such promising performance paves the way for practical uses in generating synthetic data in unmapped areas and may motivate the development of new applications of GAN in GIScience and cartography.

\begin{figure}[h]
\centering\includegraphics[width=\linewidth]{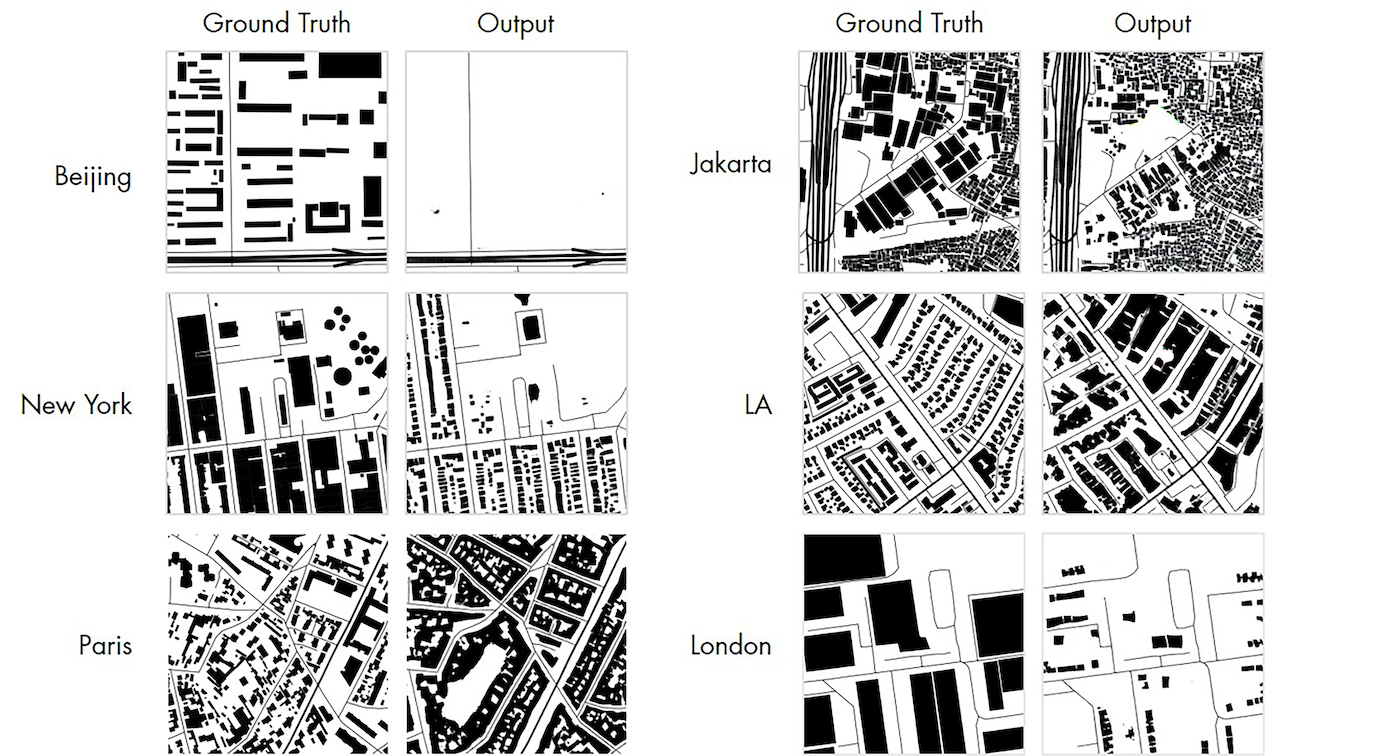}
\caption{Example of cases with inferior performance.}
\label{fig:fail}
\end{figure}

That being said, there are few limitations in the methodology that could affect the consistency and contiguity of the predicted urban texture when the tiles are stitched together. 
Figure~\ref{fig:fail} exhibits some examples of cases of the model that can be deemed as failed predictions. At zoom 16, tertiary capillaries seem to be the dominant factor in creating the generated footprints. Tiles with detailed road networks often return more compelling results. 
This effect can be seen in the failure case in Beijing where the road network has provided too little contextual information and resulted in an unsuccessful prediction. The lack of a detailed tertiary road network is a prevalent issue in the Beijing dataset which also explains Beijing's lower performance according to the FID score. It is possible that tertiary roads were not mapped completely in Beijing at the time of writing, causing some areas to suffer from the lack of information, but also indicating that our work, as it is the case with any other AI technique, is sensitive to the quality of input data.

This error due to lack of information could be reduced by using a lower zoom level such as zoom level 15 which increases the density of information of each tile. Still, it is not recommended to train the model in an area that has incomplete street or footprint data as the model could only be as good as the dataset is. 

Jakarta comes second last in terms of FID score at 122.3 due to large variations in urban texture in some tiles. In the failed example of a setting in Jakarta, we observe larger buildings being surrounded by smaller, informal settlements. This occurs for many tiles and is a unique characteristic of Jakarta. The model struggles in these areas by returning smaller settlements or vacant areas instead of larger building footprints as shown in the ground truth. In contrast to Beijing, which requires more information per tile, the extremely fine urban grain in Jakarta could benefit from a larger zoom level such as 17 so that each tile would contain more uniform typologies for interpretation.

In other cities such as New York, LA, Paris and London, there are cases where the model lacks contextual information and produces the inaccurate urban typology from the ground truth.
For example, the ground truth in the Paris sample is a suburban area while the model has produced a Haussmann-like typology typically found in Paris. Moreover, in the sample from New York, the model had generated a landed-house typology, whereas the ground truth is in fact a Townhouse typology. 
This problem is probably caused by the similarity of the street network at the specific zoom level for the different typologies in the training set. In the above-mentioned cases, the road networks outside of the City of Paris are similarly laid out as those outside of Paris while those in Uptown Manhattan are also similar to Midtown Manhattan, causing confusion in the generated typologies. 

\begin{figure}[h]
\centering\includegraphics[width=0.9\linewidth]{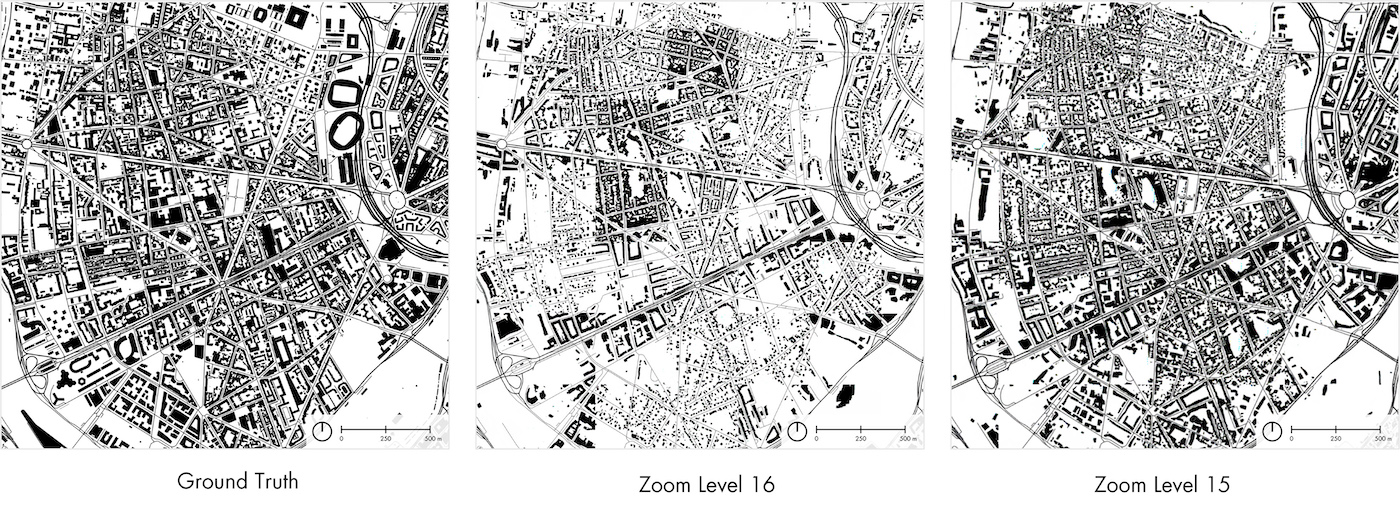}
\caption{Stitched output of Paris, compared with different scales for the same area, reasserting the influence of the zoom level on the performance of the method.}
\label{fig:Paris}
\end{figure}

As mentioned in Section~\ref{sc:exp2}, lower zoom levels capture more contextual data that helps the model to infer the correct urban typology. Thus, choosing the correct zoom level could solve the confusion between different typologies effectively. 
Figure~\ref{fig:Paris} shows that the problem of typology confusion or spatial inconsistency is prominent in Paris at zoom level 16. Most of the generated tiles are of lower density rather than the dense, courtyard-like typology in the ground truth. Using a lower zoom level of 15, the contiguity problem is improved significantly with most tiles correctly showing the proper density and texture as the ground truth.

As the successful cases in Figure~\ref{fig:exp4all} had shown, the problem of texture contiguity depends on specific urban morphologies. While zoom level 16 works well in some cities, cities such as Paris require at least zoom level 15 as its footprint typology contrasts sharply between the City of Paris and the Greater Paris area while the street networks remain similar.

Another approach to this problem is to train different models to generate the desired urban texture. For example, if the output area is within the City of Paris, a model trained only on tiles within the City of Paris would perform better than a model trained on the entire Paris metropolitan area. 

It is also important to note that the training dataset is sourced from OSM which is a volunteered geospatial dataset with heterogeneous quality.  While the quality is generally good, OSM contributors across the world may have different approaches in mapping buildings (e.g.\ level of detail and definition of buildings).
Therefore, it remains to be investigated how different mapping approaches might have an impact on the training process

We have demonstrated that a properly configured GANmapper is able to learn the predominant urban texture of a city and generate realistic and morphologically correct urban patterns in a previously unseen area in the city or another city that has a similar urban form. As long as the limitations are considered, our method opens up practical uses in generating synthetic data for a variety of applications. 

According to \citet{barrington2017world}, more than 80\% of the roads in the world are mapped in OSM (with presumably further development since their study was conducted), whereas footprints are much less complete. This approach provides an opportunity to train a model in an area that is fully mapped with both street network and building footprints, and transfer the learned patterns to an area that has street networks but missing footprint data.
It is also beneficial for locations that already have building footprints mapped but partially, having our method to supplement them and help reach full completeness.
However, in data deserts with only a few roads, a model trained on a more complete dataset might only generate buildings around the road since there is no other contextual information available.

In addition, one can also train a model on areas with little buildings and apply that to another. We assume that the street network offers sufficient information for the model to infer the density of the buildings. In fringe areas, the road networks are also likely to be sparse, consisting of a few major arteries, and the model should be able to account for these low density situations as well.

It is important to note that due to the unsupervised nature of GANs, the generator of the GAN does not actually `see' the ground truth during training. Therefore, the exact location of each predicted building polygon might not represent the correct location of the ground truth. The randomness at the building scale means that the generated data cannot be directly used in OSM in place of peer-reviewed labels from the volunteers. Rather, the models are sufficiently accurate in generating morphologically correct urban forms in different zoom levels.

One area which requires additional research is the inclusion of cartographic metrics such as number of buildings, area change and length change between the ground truth and the synthetic dataset. The current method is unable to support the calculation of these metrics as the synthetic datasets are generated in raster images. Comparison between individual building polygons using the above cartographic metrics would definitely shed more insight to the behaviour of GANs in geographic data translation and gives urban planners and geographers more confidence in the results. Perhaps an innovative vectorization algorithm such as proposed by \citet{ijgi7020041} can be applied to the results, and this could become a topic of interest for further research.

The generated building data would be useful by researchers at the urban scale.
With a trained model, planners would be able to generate morphologically correct neighbourhoods by simply providing street networks without the need of writing procedural scripts and the generated data could be used to simulate urban growth or generate design options of a new district that is consistent with the rest of the urban fabric. Alternatively, the model can also be used to transfer morphologies between cities. For example, a model trained in an area with low density buildings could be applied to a high density region for studies on the effect of building morphology on urban microclimate.
Other uses of the data in growth simulation, iterative design, and downstream machine learning on GIS data could also be explored by further research.

In addition, since the density and average shape of buildings are captured in the synthetic data, it might be possible to use the generated data for OSM quality control. OSM regions that are potentially undermapped can be compared to the result of the model trained in a similar area with a high degree of completeness. By calculating the differences in site cover and tile density between the two datasets, tiles that have significantly lesser buildings could be considered as undermapped. This method may be deployed rapidly in an area and tiles that are flagged as undermapped could receive the attention of volunteers to complete the data.

While we are optimistic about future applications of our work, we would also like to caution users on the ethics of AI generated GIS contents. Historically, the misuse of GIS tools had severe geopolitical implications \citep{war}. More recently, AI generated contents across all disciplines had caused concerns in malicious uses, and this discussion has become recently relevant in GIScience with the advent of generated `Deepfake' satellite images \citep{zhao2021}. 
As this paper has introduced an even wider application of GANs in GIS, it is vital for measures and rules to be developed to control the misuse of such algorithms in the near future.

\section{Conclusion}\label{sc:conclusion}

In this paper, we take GAN to new heights in GIScience, by introducing a new application: geographical data translation.
Leveraging on their relationship, our contribution uses more commonly found and coarse geospatial data (land use and road network) to predict less common features at the finer scale (building footprints) without data and measurements on them (e.g.\ satellite imagery or land surveying).

We developed a software, GANmapper, to translate road network data into building footprint data to demonstrate the feasibility and effectiveness of geographical data translation. We hope that this new application of GAN in GIS would introduce a new approach for digital cartography and may catalyse further investigations how can GAN be leveraged in GIScience.

As mapping features such as building footprints continues to be complex and time-consuming, they remain unmapped in much of the world.
We postulate that GANmapper could be used as a solution to create maps that approximately reveal the urban form, which despite the synthetic nature of the data, may be found valuable by various use cases such as population estimation, urban morphology, energy, and climate simulations \citep{Xu.2017,Wang.2017d7c,Yuan.2019,Wang:2020jg,Fleischmann.2021,Barbour.2019,Schug.2021,Shang.2021}.
For many of these applications, the exact geometry of each building is not essential, and such applications benefit from aggregated building data (e.g.\ total area covered by buildings), which our approach accomplishes well. 

We believe that other geographic information can be synthesised as well, as our experiments suggest that using land use maps as input produces respectable results as well.
Furthermore, our data translation approach may be used to enhance the level of detail of existing datasets.
For example, generating tertiary road networks may be possible given primary and secondary road networks as input. The foreseeable challenge is in vectorisation and scaling up in resolution.

Among other applications, we anticipate that data derived with GANmapper may serve as reference data for quality checks. For example, it may be used to detect missing buildings/built-up areas in datasets such as OpenStreetMap. For future work, we plan to explore such use cases, but also enhance the prediction workflow. For example, we plan to develop certain rules that would help cleaning or post-processing the data in tandem with GAN, such as removing buildings from locations where they are unlikely to be found (e.g.\ inside a roundabout).

\section*{Data and Codes Availability Statement}

The code that support the findings of this study are available on Github at \url{https://github.com/ualsg/GANmapper}. A version with trained checkpoints is available at \url{https://doi.org/10.6084/m9.figshare.15103128}

\section*{Acknowledgements}

We gratefully acknowledge the input data used in this research and the valuable comments by the editor and the reviewers.
We thank the members of the NUS Urban Analytics Lab for the discussions.
This research is part of the project Large-scale 3D Geospatial Data for Urban Analytics, which is supported by the National University of Singapore under the Start Up Grant R-295-000-171-133.

\section*{Notes on contributors}

Abraham Noah Wu is a research assistant at the National University of Singapore. He holds a Master Degree in Architecture from the National University of Singapore. 

\vspace*{.5cm}

Filip Biljecki is an assistant professor at the National University of Singapore and the principal investigator of the NUS Urban Analytics Lab.
He holds an MSc in Geomatics and a PhD in 3D GIS from the Delft University of Technology in the Netherlands.

\bibliographystyle{elsarticle-harv} 
\bibliography{cas-refs}

\end{document}